\documentclass{article}
\usepackage{PRIMEarxiv,amsmath,graphicx}

\usepackage{enumitem}

\usepackage{mwe} 

\usepackage{graphicx}
\usepackage{textcomp}
\usepackage{xcolor}

\usepackage{algorithmic}
\usepackage{array}
\usepackage{mdwmath}
\usepackage{mdwtab}
\usepackage{eqparbox}
\usepackage{subcaption}
\usepackage{stfloats}
\usepackage{url}
\usepackage{cite}
\usepackage{hyperref}
\usepackage{indentfirst}
\usepackage{parskip}
\usepackage{graphicx}
\usepackage{multirow}
\usepackage{tabularx}
\usepackage{booktabs} 
\usepackage{lipsum}
\usepackage{graphicx, caption}
\graphicspath{ {./images/} }

\usepackage{subfiles}

\usepackage[acronym]{glossaries}

\usepackage{amsmath}
\usepackage{ upgreek }
\usepackage[utf8]{inputenc}

\graphicspath{{Images/}{../Images/}}

\title{Temporal Feature Weaving for
Neonatal Echocardiographic Viewpoint Video Classification}

\author{
  Satchel French\\
  Electrical, Computer, and Biomedical Engineering \\
  Toronto Metropolitan Unversity \\
  \texttt{satchel.french@torontomu.ca} \\
   \And
  Faith Zhu \\
 Pediatrics \\
  Mount Sinai Hospital \\
  \texttt{Faith.Zhu@sinaihealth.ca} \\
  \And
  Amish Jain \\
 Pediatrics \\
  Mount Sinai Hospital \\
  \texttt{Amish.Jain@sinaihealth.ca} \\
  \And
  Naimul Khan\\
   Electrical, Computer, and Biomedical Engineering \\
  Toronto Metropolitan Unversity \\
  \texttt{n77khan@torontomu.ca} \\}
  
\begin{document}
  \maketitle

\begin{abstract}
Automated viewpoint classification in echocardiograms can help under-resourced clinics and hospitals in providing faster diagnosis and screening when expert technicians may not be available. We propose a novel approach towards echocardiographic viewpoint classification. We show that treating viewpoint classification as video classification rather than image classification yields advantage. We propose a CNN-GRU architecture with a novel temporal feature weaving method, which leverages both spatial and temporal information to yield a 4.33\% increase in accuracy over baseline image classification while using only four consecutive frames. The proposed approach incurs minimal computational overhead. Additionally, we publish the Neonatal Echocardiogram Dataset (NED), a professionally-annotated dataset providing sixteen viewpoints and associated echocardipgraphy videos to encourage future work and development in this field. Code available at:  \url{https://github.com/satchelfrench/NED}

\end{abstract}

\section{Introduction}

This work aims to solve the problem of neonatal echocardiogram viewpoint classification by utilizing modern machine learning algorithms.  This could broaden access to non-specialized technicians, provide improved training experiences and overall aid in the acquisition of higher quality echocardiograms. In a scan the anatomy is observed from a variety of perspectives, aptly referred to as “viewpoints”. Some viewpoints are easy to identify for a novice, but several are only separable by minor details. Additionally, viewpoints can appear different between patients, pathology's and scan environment. Prior approaches to viewpoint classification have treated it as a simple image classification problem \cite{madani1}, utilizing off-the-shelf CNN-based networks such as VGG16. Those techniques have been effective in establishing a solid baseline, but they suffer from confusion on harder to distinguish viewpoints and have greater sensitivity to input variation, leaving considerable room for improvement. Additionally, though there are solutions explored for adult echocardiograms or alternate anatomies in neonates, there is a lack of literature supporting the use case for neonatal echocardiograms. Considering the anatomy of a heart is dynamic, a particular viewpoint may appear different during an alternate position of the cardiac cycle. Prior single-image classification approaches have neglected this salient information by not considering the movement of structures within the viewpoint. We present a multi-frame approach using a gated recurrent unit (GRU) network, \cite{chungru} enabling the network to factor temporal information into its classification output. We demonstrate that this fundamental consideration of temporal-spatial relationship is a key factor in producing state of the art accuracy in viewpoint classification. key contirbutions are:

\begin{itemize}
    \item{The first of its kind, professionally labelled, open-source neonatal echocardiogram dataset featuring sixteen viewpoints, in both video and image formats.}
    \item{A modified ResNet-GRU architecture with our Temporal Feature Weaving (TFW) method which produces a 4.53\% improvement in F1-Score compared to baseline image classification.}

\end{itemize}

Our final proposed method with temporal feature weaving comes at no additional cost in model size or compute. The final model features approximately 30 million parameters, which is capable of being executed in real-time on a modern smartphone, furthering the accessibility of neonatal echocardiography.

\section{Proposed Approach}

\subsection{Dataset}

\begin{figure}[!ht]
    \begin{subfigure}[t]{0.47\textwidth}
      \includegraphics[width=\textwidth]{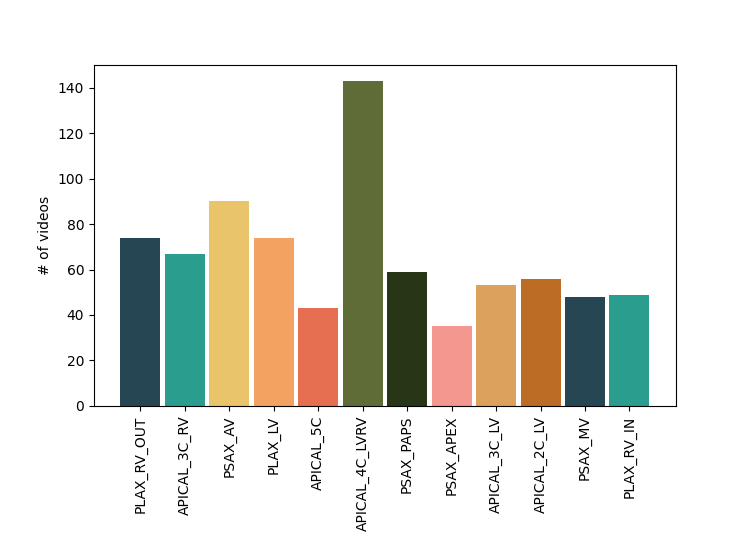}
      \caption{NED-12}
    \end{subfigure}
    \hfill
    \begin{subfigure}[t]{0.52\textwidth}
      \includegraphics[width=\textwidth]{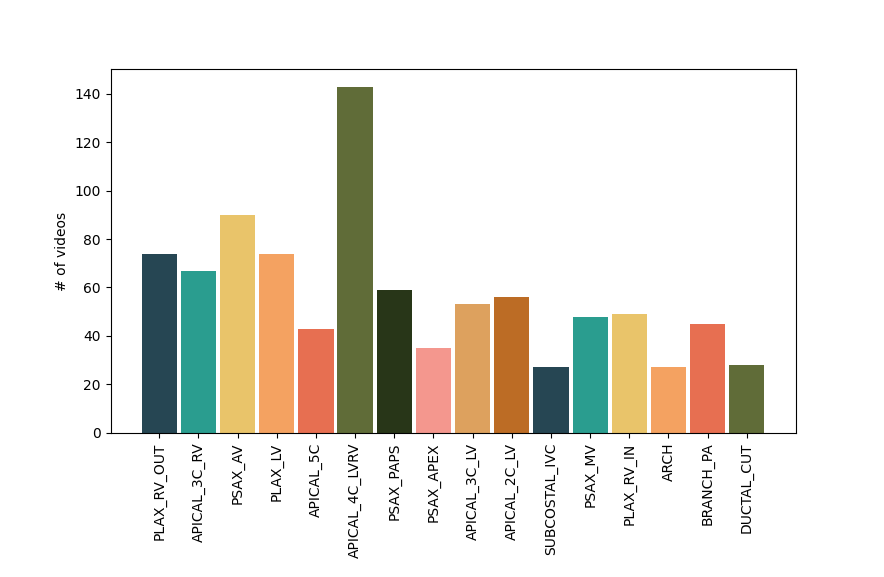}
      \caption{NED-16}
    \end{subfigure}
    \caption{Class Distribution of NED}
    \label{fig:classbalance}
\end{figure}

Our work introduces the first publicly available Neonatal Echocardiogram Dataset (NED). The dataset contains a total of 16 classes across 40 patient cases, totaling 1049 videos approximately 1 second in length. The scans are acquired organically from real patient scans, and as such contain a natural class imbalance. A subset of 12 common classes has been chosen as the primary focus of this work given their clinical relevance, scan homogeneity and sample size. These two subsets are referred to as the NED-16 and NED-12 respectively. The class balance of both datasets can be seen in Figure \ref{fig:classbalance}, where it should be highlighted that the three classes with the least samples, were removed to improve balance. A fourth class, representing the Ductal Cut viewpoint is removed as it presents with a high variation between scans and is less commonly captured. The remaining subset of 12 classes and views are selected to be consistent with a generally standard capture procedure \cite{grooves1}.

Each scan in the dataset has been labelled by a physician with expertise and familiarity in the field and individual frames are organized respectively by patient, viewpoint, and video clip. Our dataset is accompanied by open-source PyTorch data-loaders that support loading frames as consecutive or evenly spaced sequences. Additionally, it includes sequence-wide augmentation transforms for rotation, scale, shift, flips and contrast adjustment \footnote{Github link to dataset and code to be provided upon acceptance.}.

Neonatal Echocardiogram Dataset (NED) offers the following 16 classes for viewpoints: Apical-4-Chamber ($APICAL\_4C\_LVRV$), Apical-5-Chamber ($APICAL\_5C$), Apical-3-Chamber Right Ventricle ($APICAL\_3C\_RV$), Apical-3-Chamber Left Ventricle ($APICAL\_3C\_LV$), Apical-2-Chamber ($APICAL\_2C\_LV$), Parasternal Long Axis (PLAX) Left Ventricle ($PLAX\_LV$), PLAX Right Ventricle (RV) In-flow Focus ($PLAX\_RV\_IN$), PLAX Right Ventricle (RV) Out-flow focus ($PLAX\_RV\_OUT$), Parasternal Short Axis (PSAX) Apex ($PSAX\_APEX$), PSAX Papillary Muscle ($PSAX\_PAPS$), PSAX Mitral Valve ($PSAX\_MV$), PSAX Aortic Valve ($PSAX\_AV$), Pulmonary Artery Branch ($BRANCH\_PA$, NED-16 only), Ductal Cut ($DUCTAL\_CUT$, NED-16 only), Aortic Arch ($ARCH$, NED-16 only), and Subcostal Inferior Vena Cava ($SUBCOSTAL\_IVC$, NED-16 only).



Patient demographics available in Table \ref{table:demographics} highlights the group statistic of dataset participants. Notable statistics include a moderate skew toward male patients at 61.8\%, and median gestational age at time of scan of 32.15 weeks.
\begin{table}[htbp]
\setlength\tabcolsep{0pt} 
\caption{Patient Demographics}
\begin{tabular*} {\columnwidth}{@{\extracolsep{\fill}} ll *{2}{r} }
   \toprule
   \multicolumn{2}{l} {Metric} & Value (std)\\
   \midrule
        \multicolumn{2}{l}{Patient Count} & 34 \\
        \multicolumn{2}{l}{Gestational age at birth (weeks)} & 25.87 (1.84) \\
        \multicolumn{2}{l}{Gestational age at scan (weeks)} & 32.15 (5.2) \\
        \multicolumn{2}{l}{Age at scan (days)} & 46 (40) \\
        \multicolumn{2}{l}{Male sex (\%)} & 21 (61.8) \\
        \multicolumn{2}{l}{Weight at scan (kg)} & 1.67 (1.01) \\
        \multicolumn{2}{l}{Systolic blood pressure (mmHg)} & 72 (13) \\
        \multicolumn{2}{l}{Diastolic blood pressure (mmHg)} & 42 (10)) \\

   \bottomrule
\end{tabular*}
\label{table:demographics}
\end{table}

\subsection{Model Architecture}

The foundational ResNet-50 by \textit{He et al.} \cite{he2015resnet} is used for both the image classifier and backbone of GRU based models. Given efficacy of VGG style networks for feature extraction \cite{vgg} \cite{madani1}, ResNet is expected to provide same or better performance with fewer layers given its identity mapping via skip connections. In the context of GRU networks, ResNet enables the selection of both high-level and low-level features across time, providing richer feature combinations.

To compare image classifier performance, mean-voting of class probabilities across several frames was employed. It produces marginal improvement overall, and performs better on spaced sequences of frames, compared to a consecutive sequence. This is expected behaviour as given the model's conviction in its predictions, improvement would be seen only in cases where classes have a similar confidence, and would be better differentiated at alternate points during the cardiac cycle where anatomy is more dissimilar or imaging is less noisy.

\subsubsection{Self-Attention Methods}

To better understand the efficacy of this approach, a benchmark using an attention mechanism \cite{vaswani2023attentionneed} is employed. This is a sensible approach given the success of attention models for sequence learning in relevant tasks \cite{arnab2021vivitvideovisiontransformer} \cite{2stream}. Several architectures were tested including the use of GRU with attention, but the best results were ultimately produced by splitting the 2048 length CNN feature outputs of four frames into a 512x16 sequence. This was fed into four attention heads, and the attended sequence was fed into an fully connected layer of 512x2048, before going to the 2048x12 class output layer. 

\subsection{Temporal Feature Weaving}

\begin{figure}[!hb]
\centering
\includegraphics[width=\linewidth]{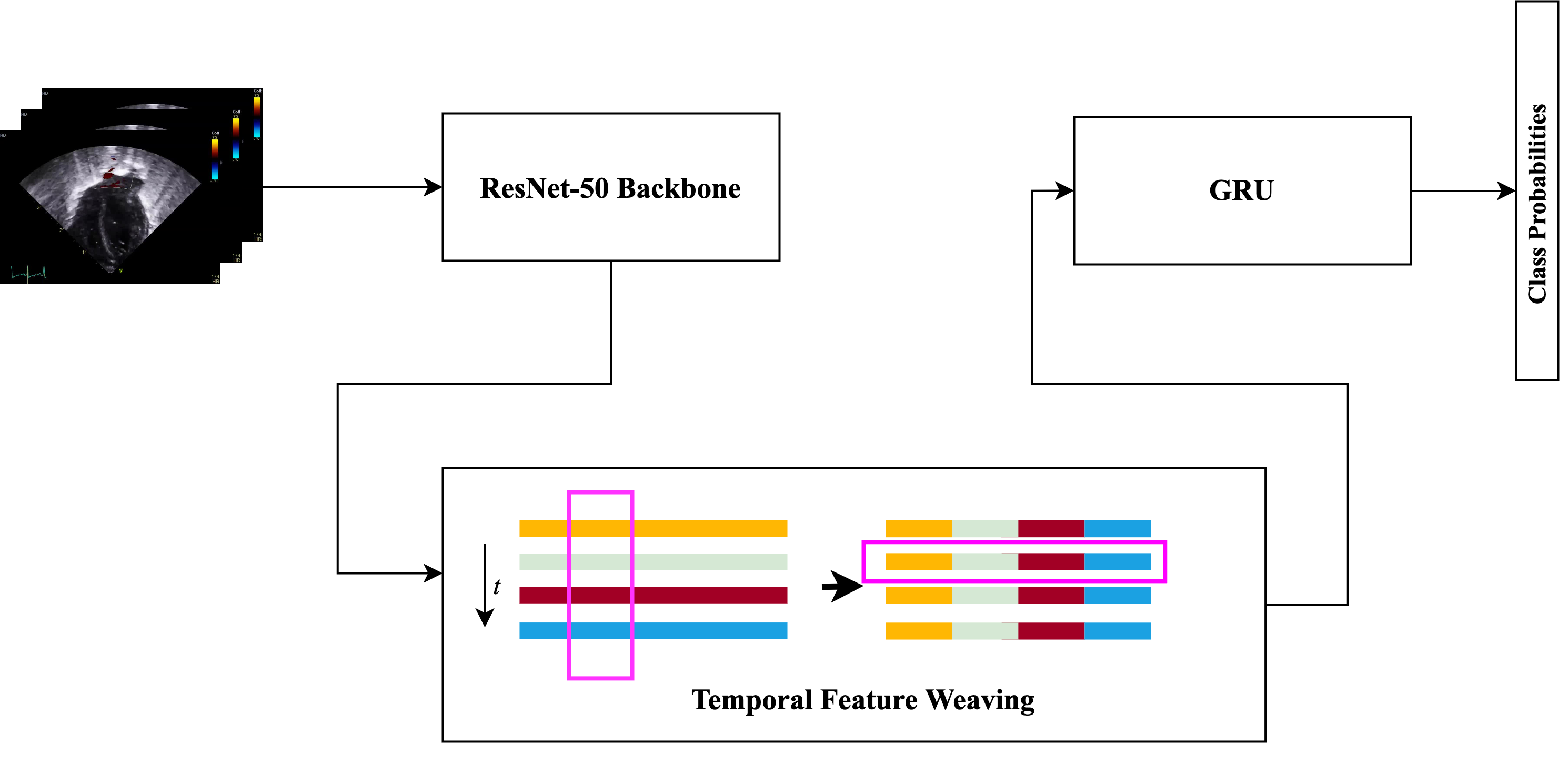}
\caption{ResNet-50-GRU-Temporal-Weave Architecture}
\label{fig:tfw}
\end{figure}

To capture more meaningful temporal-spatial relationships we introduce the concept of temporal weaving. Temporal weaving divides the flattened feature vectors of each frame $X_n$ into $K$ segments (Equation \ref{1}), and reconstructs a new feature vector, $W_k$, by concatenating the respective chunks, $C$, between frames as seen in (Equation \ref{2}). Visually this is captured in the architecture Figure \ref{fig:tfw}. In this paper we employ a simplistic approach to temporal weaving where $K=N$, meaning the weaved feature vector $W$, represents the same chunk evenly across time.

\begin{equation}\label{1}
    X_n = [C_{n1}, C_{n2}, C_{n3}, ... ,C_{n(K-1)}, C_{nK}]
\end{equation}
\begin{equation}\label{2}
    W_k = [C_{1k}, C_{2k}, C_{3k}, ... , C_{(N-1)k}, C_{Nk}]
\end{equation}

Our approach produces $N$ weaved feature vectors, which are given as a sequence input to the GRU. Predictions are made at at each time-step $t$, with final class probabilities being taken from the final time-step. 

Conceptually, this encourages the model to give attention to the same patch of the image across time, highlighting temporal information. Weaving the features this way provides a temporal signature in each sequence input, which can easily be obtained by contrasting chunks from different frames. This is thought to improve upon frame-sequence input as the ResNet backbone produces solely spatial features which are effectively compressed into smaller subspaces at each time-step in the GRU's hidden layer, making it unlikely that the minor differences across time persist as salient features. By weaving the features, the network is forced to make predictions on a single spatial-temporal feature patch, which when viewed as independent events, are implicitly chained together by the persistence of information in an RNN hidden layer. The end-of-The weaved features may be thought of as a special case where more relevant information persists to the end-of-time prediction.

\section{Experiments}
\subsection{Training Parameters and Baseline Models}

Data augmentation proved to be essential in producing quality results without over fitting the dataset. However, augmentation for echocardiograms requires more careful consideration to ensure that the transforms produced clinically viable samples. Our research indicated that a random rotation of [0, 25] degrees and auto contrasting yielded the best results. These findings are sensible from a clinical perspective and are consistent with literature \cite{tupper2024analyzingdataaugmentationmedical}.

As a baseline image classifier architecture, ResNet-50 \cite{he2015resnet} is trained from scratch on NED-12 (pre-training pm ImageNet provided no benefit). The model was trained for 50 epochs using cyclical learning rate scheduling with a maximum learning rate of (lr=0.08) and batch size of (N=16). It uses a modified gradient descent algorithm, with a momentum of (m=0.75), L2 regularization of (d=5e-4), and dropout of (p=0.6). Finally, samples are generated by loading a single frame from random patients scans, resized to 232x232, cropped to 224x224, randomly augmented, and sampled with no-replacement for the epoch. 

For the video classification model there was an obvious need to provide interleaved frames. While the model performs well with several sequential frames, by processing spaced frames it encodes temporal information that may be unique to a viewpoint. Additionally, viewpoints may appear dramatically different throughout the cardiac cycle, with some parts of the anatomy moving slowly early in the scan, and then quickly moving during another segment.

When training the GRU based model, initial results failed to make meaningful improvements as training continued, likely due to the vanishing gradient problem recurrent neural networks suffer from. Although it’s worth noting superior training performance was observed with GRU networks compared to LSTM, the reduction in loss was minimal. To better address these concerns, weights from the trained resnet-50 classifier were used, with outputs from the second-last fully connected layer being flattened into a feature vector of 2048 length. These features are generated for each frame in the sequence and weaved temporally before being fed into the model as a sequence of feature vectors. The GRU based model was also trained using cyclic learning rate scheduling with a maximum learning rate of (lr=0.003) and a batch size of (N=16) for 60 epochs. It also used Adam for gradient descent, with a momentum of (m=0.75), L2 regularization of (d=1e-3) and dropout (p=0.5).

Models were initially evaluated on a random 90/10 train test split, and cross validated with (K=5) folds. Folds are split patient wise to mitigate data leakage from any consistencies in a patients scan. Metrics are calculated during each fold and finally aggregated into accuracy, precision and recall scores.

\subsection{Results} 

Table \ref{table:results} summarizes the results of experiments on the NED-12, including the proposed ResNet-GRU-TFW network, as well as comparisons to attention-based and mean-voting approaches. Our proposed solution of ResNet-GRU with Temporal Feature Weaving (TFW), produces an accuracy of 93.8\%, and an F1-Score of 93.7\% when evaluated on spaced frames outperforming all other models.

\begin{table}[htbp]
\caption{Results on NED-12}
\setlength\tabcolsep{2pt} 

\begin{tabularx}{\linewidth}{@{\extracolsep{\fill}} X *{4}{r} r r r r}
   \toprule
   Model & Accuracy & Precision & Recall & F1 \\
   \midrule
   ResNet-50-Image-Classifer & 89.55 & 88.99 & 89.55 & 89.20 \\
   Mean-Voting-4-Consecutive & 89.73 & 90.66 & 89.73 & 90.01 \\
   Mean-Voting-4-Spaced & 89.99 & 90.95 & 89.99 & 90.27 \\
   ResNet-GRU-4-Consecutive & 88.46 & 88.84 & 88.46 & 88.44 \\
   ResNet-Attention-512-Consecutive & 92.13 & 92.14 & 91.36 & 91.19 \\
   \cmidrule{1-5}
   \textbf{(Proposed) ResNet-GRU-TFW-4-Spaced} & \textbf{92.55} & \textbf{92.72} & \textbf{92.55} & \textbf{92.51} \\
   \textbf{(Proposed) ResNet-GRU-TFW-4-Consecutive} & \textbf{93.88} & \textbf{93.65} & \textbf{93.88} & \textbf{93.73} \\
   \bottomrule
\end{tabularx}

\label{table:results}
\end{table}

\begin{figure}[!ht]
    \centering
    \begin{subfigure}[t]{0.45\linewidth}
        \includegraphics[width=\linewidth]{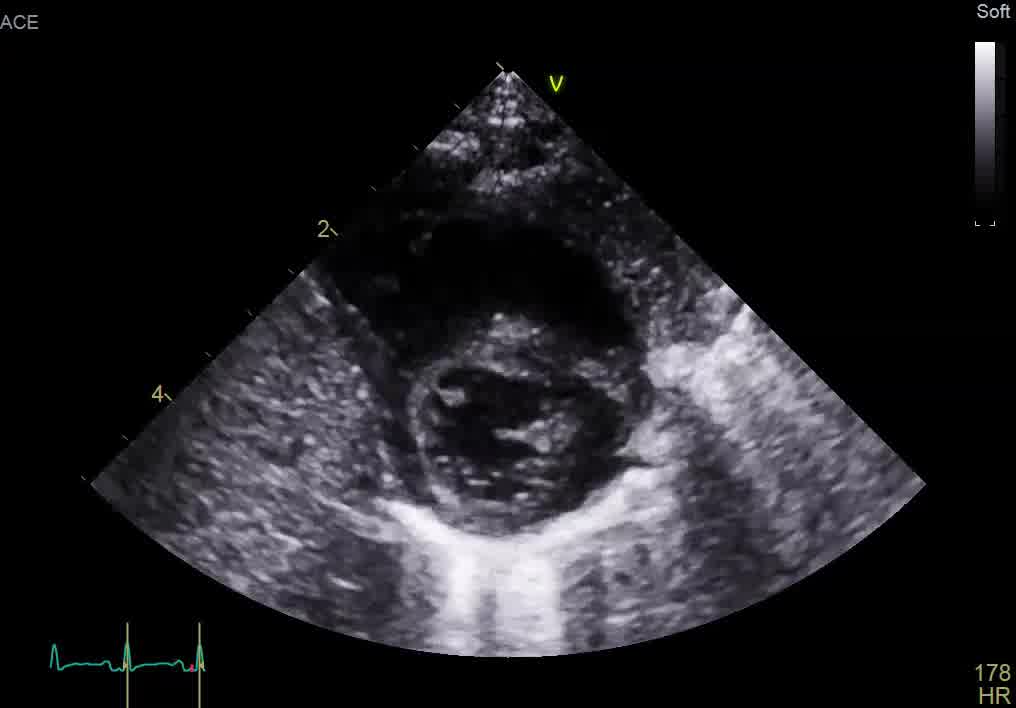}
        \caption{}
    \end{subfigure}
    \hfill
    \begin{subfigure}[t]{0.45\linewidth}
        \includegraphics[width=\linewidth]{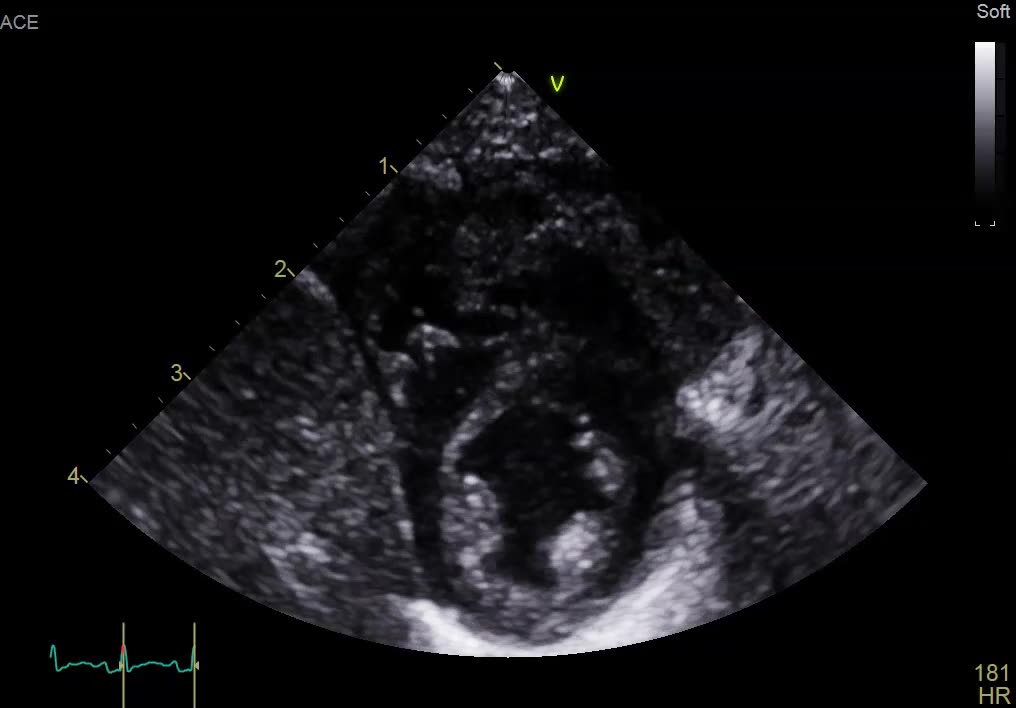}
        \caption{}
    \end{subfigure}

    \caption{(a) Mitral-valve ($PSAX\_MV$) (b) Papillary muscle ($PSAX\_PAPS$)}
    \label{fig:mvus}
\end{figure}

\section{Discussion}
\label{chap:dis}

As seen in Table \ref{table:results}, there is a general progression of improvement in accuracy from single frame to multi-frame sequential and spaced configurations. Spaced-frame voting offers a better perspective throughout the cardiac cycle, where structures may appear meaningfully different. The ResNet-GRU architecture without temporal feature weaving performs worse across all metrics, showing a 0.76\% decrease in F1 Score compared to the baseline. This reduced accuracy may stem from the model's inability to memorize subtle differences between sequential frames, focusing instead on higher-level features. Recurrent neural networks compress input features by selecting which sequence elements to remember, potentially causing a lossy compression where certain discriminants are lost. The GRU may no longer access high-definition spatial features from the ResNet architecture, relying instead on high-level semantic features that underperform the baseline.

Substantial improvement is observed with the CNN-Attention approach, and weaving attended features made no difference as expected. This suggests the attention mechanism effectively identifies relevant patches for comparison across time, achieving the intended result. However, it still underperforms the TFW method by 2.4\% in F1-Score. Additionally, multi-head attention adds training complexity and an extra 1 million parameters, whereas TFW introduces no such overhead.

Introducing TFW improves the F1 score by 3.31\% and 4.53\% for spaced and consecutive models, respectively. These gains over mean voting approaches support the hypothesis that considering the temporal signature of individual viewpoints enhances classification. Notably, the sequential case outperforms the spaced case with temporal feature weaving, contrary to the mean-voting scenario. This indicates an alternate prediction mechanism, where the sequential case better captures structural movements, suggesting that a unique temporal signature contributes to the model's predictions.

This effect is most evident when differentiating between static views of the parasternal short axis (PSAX) mitral valve and papillary muscle viewpoints. As shown in Figure \ref{fig:mvus}, the viewpoints appear remarkably similar, differing mainly in a cropped focus on the papillary muscle and a clearer view of the mitral valve. The primary discriminator is the presence of the mitral valve, or "Fish Lips." However, since the valve is dynamic, it is not consistently identifiable by this feature alone from a single frame.

\section{Conclusion}

We propose a novel temporal feature weaving (TFW) approach to classify neonatal ultrasound videos into different viewpoints. Our work shows that tracking features across time adds value to viewpoint classification, which has not been explored in literature thoroughly. We also provide a public neonatal viewpoint classification dataset. Future work will focus on addressing limitations such as dataset size, using multiple labelers, and exploring other TFW approaches. 

\section{Compliance with ethical standards}
This study was performed in line with the principles of the Declaration of Helsinki. Approval was granted by the Research Ethics Board of the Mount Sinai Hospital (19-0065-E) and Toronto Metropolitan University (2019-389).
\section{Acknowledgements}
This work was supported by the Natural Sciences and Engineering Research Council of Canada (Alliance Grant 546302-19
and Discovery Grant 2020-05471).
\bibliographystyle{IEEEbib}
\bibliography{ref}
\end{document}